\documentclass[10pt,twocolumn]{article}

\usepackage[margin=0.75in]{geometry}
\usepackage{amsmath,amssymb,amsfonts}
\usepackage{graphicx}
\usepackage{booktabs}
\usepackage{cite}
\usepackage{hyperref}
\usepackage{caption}
\usepackage{textcomp}
\usepackage{algorithm}
\usepackage{algpseudocode}
\usepackage{multicol}
\usepackage{multirow}

\title{Design of a hybrid tendon-linkage finger with coupled and adaptive motion}

\author{
    Hyojae Kang$^{1}$, Giuo Seo$^{1}$, Jaeyoung Lee$^{1}$, Yeong Jae Park$^{1}$, Jongwoo Park$^{1}$, \\ Hyunmin Do$^{1}$, So-Hee Park$^{1}$, Dongil Park$^{1,2,*}$, and Jeongdo Ahn$^{1,*}$ \\
    $^{1}$Advanced Robotics Research Center, Korea Institute of Machinery \& Materials (KIMM) \\
    $^{2}$Mechanical Engineering, University of Science \& Technology (UST) \\
    $^{*}$Co-Corresponding author
}

\begin{document}
\date{}
\maketitle

\thispagestyle{plain}
\footnotetext{This manuscript has been submitted for possible publication.}

\begin{abstract}
Dexterous robotic fingers require coordinated joint motion and adaptability to object contact, while independently actuating multiple joints increases the complexity of the transmission mechanism. This paper presents a hybrid tendon--linkage robotic finger with four mechanical degrees of freedom driven by three actuators. The MCP abduction/adduction joint is driven through a bevel gear, while MCP and PIP flexion/extension are driven by separate bidirectional tendon transmissions. The DIP joint is not independently actuated but is coupled to the PIP joint through a four-bar linkage with a torsional spring and mechanical stopper. The tendon routing also introduces mechanical coupling between the MCP and PIP joints. Kinematic analysis was conducted to describe the coupled and adaptive motions, and experiments were performed to evaluate the joint motion characteristics. ArUco-based measurements showed that the coupled motion closely matched the theoretical prediction, whereas the adaptive motion was partially limited by interference between linkage components. The fabricated finger demonstrated different contact postures with cylindrical, cuboid, and irregularly shaped objects, and the mean peak fingertip force was 4.56 N. The proposed mechanism provides both coupled motion and contact-dependent adaptation without requiring an additional actuator for the DIP joint.
\end{abstract}

\section{Introduction}

Robotic fingers for grasping require a balance among compact size, output force, adaptability, and controllability. Independently actuated dexterous hands provide high joint controllability but generally require many actuators and transmission components \cite{c9,c90,c95}. Underactuated mechanisms reduce this requirement by allowing part of the finger configuration to be determined mechanically \cite{laliberte1998simulation,tavakoli2015underactuated,c14,c91,c88}. Underactuated mechanisms can be implemented using mechanical coupling or stopper with elastic components such as springs. As more mechanical components are incorporated into the finger, its size and weight increase. To reduce the size and weight of the finger, it is advantageous to locate the motors and reduction mechanisms outside the finger. Tendon transmission is widely used to locate actuators outside the moving finger links and transmit their motion remotely \cite{inouye2012novel,kurita2011human,c15,c92,min2025antagonistic}. 

Mechanical coupling provides a practical means of coordinating multiple joint motions without independently actuating every joint \cite{laliberte1998simulation,tavakoli2015underactuated}. Rigid linkages are particularly useful because their input--output relationship is defined by geometry and can vary with configuration \cite{c1,yoon2016underactuated,su2023sau,cheng2025design}. A purely fixed coupling, however, may restrict additional distal motion once one phalanx is constrained by contact. Self-adaptive mechanisms address this limitation by introducing passive mobility or elastic elements that allow the joint relationship to change after contact \cite{birglen2004kinetostatic,birglen2010flapping,qiao2019self}. Coupled motion and adaptive motion therefore serve complementary roles: the former defines the nominal free-space trajectory, whereas the latter provides additional configuration change under contact constraints.

Coupled and self-adaptive behaviors have been implemented using gears, belts, linkages, springs, and hybrid transmission mechanisms \cite{li2010study,sun2012novel,zhang2013two,li2012development,li2014coupled,li2014indirectly,liu2019coupled}. Hybrid mechanisms combining tendon transmission, linkages, and elastic elements have also been investigated \cite{c20,lee2023design,zhang2023design,li2026design}. For remotely actuated serial fingers, however, independently actuating multiple distal joints can increase the number of tendon paths crossing proximal joint regions and thereby increase routing complexity and transmission losses \cite{kim2025dual}. This motivates an architecture in which active remote actuation, nominal distal coupling, and contact-dependent adaptation are assigned to distinct mechanical elements.

\begin{table*}[!t]
\centering
\caption{Comparison of representative coupled and adaptive robotic finger mechanisms.}
\label{tab:coupled_adaptive}
\begin{tabular}{p{2.5cm} p{4.7cm} p{4.0cm} p{4.0cm}}
\toprule
\textbf{Study} &
\textbf{Transmission mechanism} &
\textbf{Free-space behavior} &
\textbf{Adaptive behavior} \\
\midrule

Li and Zhang \cite{li2010study} &
Gear--rack + pulley--belt + spring &
Coupled motion &
Self-adaptive motion after contact \\

Li et al. \cite{li2014coupled} &
Gear + pulley--belt + spring &
Coupled motion &
Directly self-adaptive motion \\

Liu et al. \cite{liu2019coupled} &
Double-linkage--slider &
Coupled motion &
Indirectly self-adaptive motion \\

Yoon et al. \cite{yoon2016underactuated} &
Five-bar + four-bar linkages + torsion springs + stoppers &
Mechanically coordinated three-joint motion &
Contact-induced passive reconfiguration \\

Su et al. \cite{su2023sau} &
Flexible belt + four-bar linkage + springs &
Coupled middle--distal motion &
Belt- and spring-mediated adaptation \\

Lee and Chang \cite{lee2023design} &
Tendon actuation + four-bar linkage + torsion springs &
Actuated MCP/PIP motion with coupled PIP--DIP motion &
Tendon-driven self-adaptive mode \\

Cheng et al. \cite{cheng2025design} &
Linkage + prismatic pairs + compression spring &
Three-joint coupled motion &
Adaptive reconfiguration after contact \\

\textbf{Proposed finger} &
\textbf{Bidirectional remote tendon transmission + bevel gear + four-bar linkage + torsion springs} &
\textbf{Active MCP A/A and MCP/PIP F/E with coupled PIP--DIP motion} &
\textbf{Contact-triggered deviation of the distal linkage} \\
\bottomrule
\end{tabular}
\end{table*}

Therefore, this study presents a hybrid tendon--linkage robotic finger without embedding motors in the moving phalangeal links. The finger provides four mechanical degrees of freedom (DoFs) using three actuators: MCP abduction/adduction (A/A) is transmitted through a bevel-gear mechanism, while MCP and PIP flexion/extension (F/E) are driven by separate bidirectional remote tendon inputs. The PIP tendon is routed through the MCP region, introducing a mechanically defined interaction between MCP and PIP motion. The DIP joint has no dedicated actuator and instead follows a nominal PIP--DIP coupling defined by a local four-bar linkage. A torsional elastic element and mechanical stopper permit the distal linkage to deviate from this nominal relationship when the PIP motion is constrained by contact. The configuration-dependent coupled and adaptive motions are investigated through kinematic analysis and experimental validation.

The main contributions are:
\begin{enumerate}
    \item Design of a 4-DoF underactuated robotic finger with hybrid tendon--linkage remote actuation, combining multiple transmission mechanisms without embedding actuators in the moving finger links.
    \item Implementation and kinematic characterization of stopper-constrained PIP--DIP coupling and contact-triggered adaptive motion, including the prediction of configuration-dependent maximum adaptive rotation and experimental validation using ArUco measurements.
\end{enumerate}

The remainder of this paper is organized as follows. Section~II reviews representative coupled and adaptive robotic finger mechanisms. Section~III describes the mechanical design and tendon-routing architecture of the proposed finger. Section~IV presents the kinematic analysis of the coupled and adaptive motions. Section~V experimentally evaluates the coupled motion, contact-triggered adaptive behavior, fingertip force, and object-contact performance. Section~VI concludes the paper.

\section{Related Work}

Underactuated fingers employ fewer independent actuation inputs than DoFs, allowing joint configurations to depend on the transmission, passive elements, and environmental constraints \cite{laliberte1998simulation,tavakoli2015underactuated,birglen2004kinetostatic}. In coupled motion, multiple joints follow a mechanically prescribed relationship during unconstrained motion, whereas self-adaptive mechanisms introduce passive mobility that allows this relationship to change after contact \cite{birglen2010flapping,qiao2019self}. Springs and mechanical stoppers can establish the nominal configuration while permitting additional passive motion under external constraints \cite{aukes2011varying}.

Table~\ref{tab:coupled_adaptive} summarizes the transmission mechanisms, free-space motion, and adaptive motion of representative coupled and adaptive robotic fingers. These robotic fingers have been realized using different types of transmission mechanisms. Linkage-based mechanisms have also combined prescribed free-space coordination with contact-induced passive reconfiguration using torsional or compression springs and mechanical constraints \cite{yoon2016underactuated,su2023sau,cheng2025design}. Hybrid tendon--linkage mechanisms further combine remote actuation with rigid coupling and elastic elements \cite{c20,lee2023design,zhang2023design,li2026design}.

The proposed finger differs from the representative mechanisms in Table~\ref{tab:coupled_adaptive} by integrating bidirectional remote tendon actuation, MCP--PIP mechanical coupling, and stopper-constrained PIP--DIP coupling with contact-dependent adaptation. This arrangement enables active MCP and PIP control and coupled--adaptive DIP motion without an additional distal actuator.

\section{Mechanical Design}

The proposed robotic finger provides four DoFs using three actuators. The MCP A/A motion is independently actuated through a bevel-gear transmission, whereas the MCP and PIP F/E motions are independently driven through tendon transmissions with remotely located actuators. The DIP joint has no dedicated actuator and is instead coordinated with the PIP joint through the distal linkage.

\begin{figure*}
    \centering
    \includegraphics[width=0.99\textwidth]{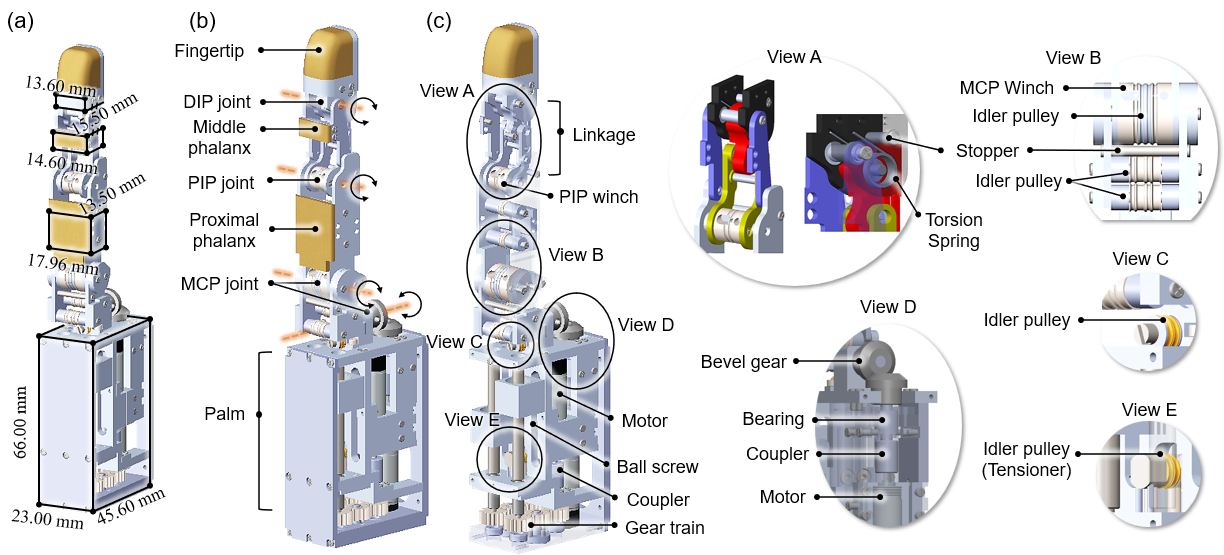}
    \caption{Overall structure of the proposed robotic finger: (a) specification, (b) joint and phalanx configuration, and (c) internal transmission components with detailed views.}
    \label{fig1}
\end{figure*}

Figure~\ref{fig1} shows the overall structure of the proposed finger. As shown in Fig.~\ref{fig1}(a), the finger width decreases toward the distal end, measuring 17.96, 14.60, and 13.60 mm at the proximal phalanx, middle phalanx, and fingertip, respectively. The thickness is 13.50 mm at the proximal phalanx and 15.50 mm at both the middle phalanx and fingertip. The lower housing containing the actuation components measures 66.00 mm in height, 23.00 mm in width, and 45.60 mm in thickness.

Figure~\ref{fig1}(b) shows the arrangement of the finger joints and phalanges. The lower housing corresponds to the palm-side region in which the actuation components are concentrated. Two rotational axes are located at the MCP region, corresponding to MCP A/A and MCP F/E motions. These two axes are spatially offset rather than intersecting at a single point. The PIP and DIP joints provide the remaining F/E motions along the serial finger chain. The contact surface of the finger is formed by the proximal phalanx, middle phalanx, and fingertip.

Figure~\ref{fig1}(c) shows the internal transmission components. Near the PIP joint, the PIP winch is driven by the tendon and transmits motion to the coupled--adaptive linkage. View A shows the detailed structure of this linkage, which consists of four links. Two intermediate links are connected by a torsion spring, and their relative rotation is restricted by a mechanical stopper during coupled motion. The stopper maintains the nominal PIP--DIP coupled motion. When contact constrains the PIP joint, further actuation allows relative rotation between the two links, producing the adaptive motion.

Below the PIP winch, idler pulleys guide the tendon paths toward the proximal actuation components. View B shows the MCP F/E transmission, including the MCP winch and the intermediate idler pulleys required to route the PIP tendons through the MCP region. A mechanical stopper limits the MCP F/E range of motion, while additional idler pulleys define the tendon paths through the finger. View C shows an additional routing pulley used to redirect the tendons toward the palm-side actuation unit.

The MCP A/A transmission is detailed in View D. Motor rotation is transferred through a coupler and bearing assembly to the input bevel gear, while the mating bevel gear forms the rotational axis of the MCP A/A joint. This arrangement allows the corresponding motor to remain outside the moving finger links while transmitting rotational motion to the MCP joint.

The actuators for driving the tendon are located in the palm-side housing. Each actuator uses a motor, gear train, and ball screw to generate linear tendon displacement. A tendon-fixing block is attached to the ball-screw nut, and motor rotation is transmitted to the ball screw through the gear train. After tendon routing, the adjustable idler pulley shown in View E is used for coarse tension adjustment, while the tendon fixation points at the winches provide additional fine adjustment.

The MCP and PIP F/E joints are each driven by a pair of tendons routed in opposing directions. This bidirectional tendon arrangement eliminates the need for elastic return elements at the actively driven MCP and PIP joints and allows both flexion and extension to be actively commanded.

\begin{figure}
    \centering
    \includegraphics[width=0.99\columnwidth]{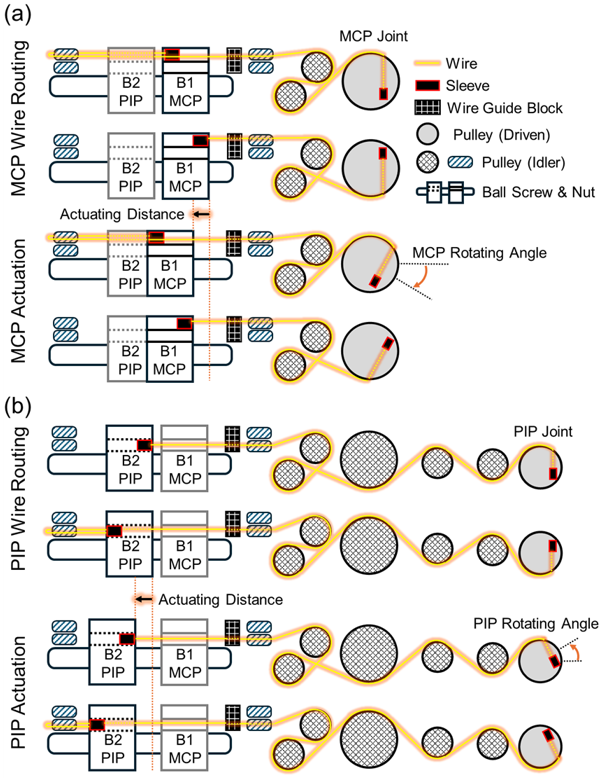}
    \caption{Tendon-routing paths and joint rotations induced by ball-screw actuation: (a) MCP F/E transmission and (b) PIP F/E transmission.}
    \label{fig2}
\end{figure}

The tendon-routing paths from the ball-screw actuation units to the MCP (B1) and PIP (B2) joints are illustrated in Fig.~\ref{fig2}. Each tendon guide block provides two independent routing paths. The path closer to the ball-screw thread is connected to the PIP driven pulley, whereas the outer path is connected to the MCP driven pulley. The idler pulleys are arranged so that the tendons remain guided along the prescribed paths throughout the joint motion and do not disengage from the pulley system.

Figure~\ref{fig2}(a) illustrates the tendon-routing paths for MCP actuation. Two tendons originate from the ball-screw nut block. One tendon passes through the wire guide block and an idler pulley before reaching the MCP driven pulley (winch). The other tendon first passes around an idler pulley located on the opposite side of the wire guide block, then follows the path through the wire guide block and another idler pulley to the MCP winch. Both ends of each tendon are secured using sleeves. When the ball-screw nut block is displaced away from the wire guide block, the MCP joint rotates as illustrated in Fig.~\ref{fig2}(a).

Figure~\ref{fig2}(b) illustrates the tendon-routing paths for PIP actuation. The two tendons connected to the PIP joint follow a similar routing arrangement near the ball-screw actuation unit. Both tendons pass around idler pulleys coaxial with the MCP driven pulley and are then routed through additional idler pulleys before reaching the PIP driven pulley. When the ball-screw nut block is displaced away from the wire guide block, the PIP joint rotates in the opposite direction to the MCP joint, as illustrated in Fig.~\ref{fig2}(b).

Because the PIP tendon path passes through the MCP region, MCP rotation changes the effective tendon length transmitted toward the PIP winch. Consequently, MCP flexion produces a mechanically induced contribution to PIP motion in addition to the independently commanded PIP input.

The actuation wires are 0.45 mm-diameter 7×19 stranded cables made of SUS304 and are mechanically terminated using crimped sleeves and soldered end fittings. The sleeves are retained at the winches and can be adjusted using set screws to increase tendon tension. Together with the adjustable idler pulley shown in View E, this arrangement enables tendon pretension to be set after assembly. The detailed tendon-tensioning mechanism and the ball-screw-based tendon actuation principle follow the authors' previous design in \cite{c99}.


\section{Kinematic Analysis}\label{sec:KA}

The planar flexion mechanism comprises two remotely actuated tendon inputs and a passive relative motion in the distal linkage. MCP F/E actuation rotates the MCP joint and additionally contributes to PIP motion through tendon routing, whereas the PIP tendon input acts on the distal transmission. The DIP joint has no independent actuator; its rotation arises from nominal PIP--DIP coupling and from passive adaptation following contact. The tendon-length variation caused by MCP A/A motion is negligible and is excluded from the planar analysis. The tendon inputs and joint coordinates are defined as

\begin{equation}
q_M, \quad q_P, \quad \boldsymbol{\theta}=
\begin{bmatrix}\theta_2&\theta_3&\theta_4\end{bmatrix}^{T},
\label{eq:coordinates}
\end{equation}

\noindent where $q_M$ and $q_P$ are the linear tendon displacements associated with MCP and PIP actuation, respectively. The passive motion of the distal linkage is defined separately below rather than being treated as an additional tendon displacement.

\begin{figure}
    \centering
    \includegraphics[width=0.99\columnwidth]{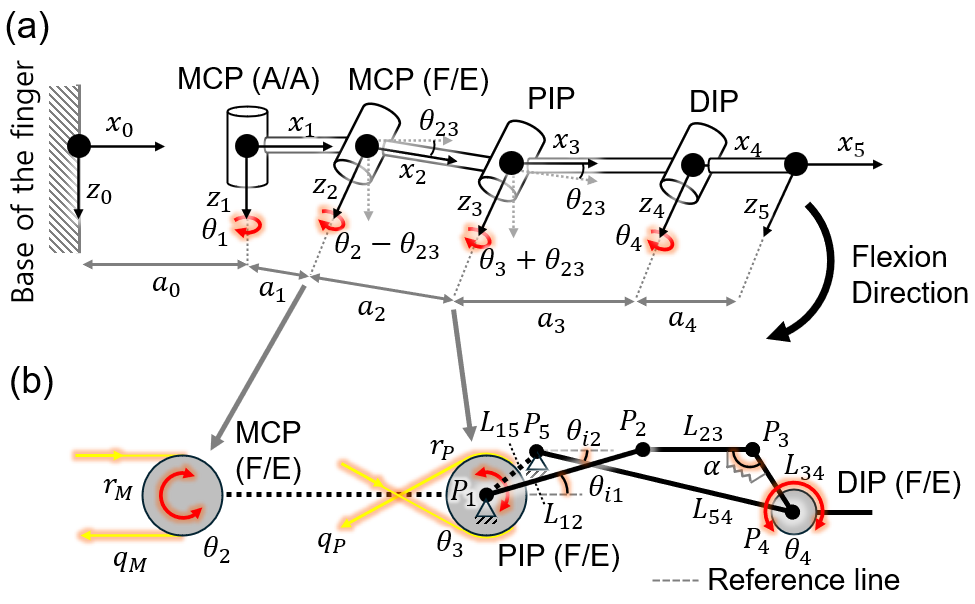}
    \caption{(a) Serial kinematic chain and coordinate frames, and (b) planar flexion transmission showing the MCP and PIP tendon displacements $q_M$ and $q_P$ and the distal coupled--adaptive linkage.}
    \label{fig3}
\end{figure}

Figure~\ref{fig3}(a) shows the serial kinematic chain of the finger. The MCP A/A joint is represented by $\theta_1$, while $\theta_2$, $\theta_3$, and $\theta_4$ denote the relative F/E rotations of the MCP, PIP, and DIP joints, respectively. The coordinate frames are assigned along the serial chain for forward-kinematic calculation, with the link dimensions $a_0$--$a_4$ and the rotational offset $\theta_{23}$ defined as shown.

The positive direction of ball-screw actuation is defined as the displacement toward the wire guide block for both actuation units. According to the coordinate frames in Fig.~\ref{fig3}(a), positive MCP actuation $q_M$ produces counterclockwise rotation of $\theta_2$ about the $z_2$-axis, whereas positive PIP actuation $q_P$ produces clockwise rotation of $\theta_3$ about the $z_3$-axis.

Figure~\ref{fig3}(b) illustrates the planar flexion transmission. The MCP tendon displacement $q_M$ generates MCP flexion and simultaneously changes the effective length of the tendon routed toward the PIP joint, thereby inducing an additional PIP rotation. The PIP tendon displacement $q_P$ supplies the other active input. At the distal side, the PIP and DIP joints are connected through the linkage composed of $L_{12}$, $L_{23}$, $L_{34}$, and $L_{54}$.

The MCP flexion angle is related to its tendon displacement by

\begin{equation}
\theta_2=\frac{q_M}{r_M},
\label{eq:mcp_angle}
\end{equation}

\noindent where $r_M$ is the effective winding radius defined by the tendon centerline at the MCP transmission. MCP rotation also changes the effective tendon displacement at the PIP transmission. With $r_P$ denoting the effective PIP winding radius, the induced PIP rotation during nominal coupled motion is

\begin{equation}
\theta_3^{M}=\frac{r_M}{r_P}\theta_2,
\label{eq:mcp_pip_coupling}
\end{equation}

\noindent where $\theta_3^{M}$ is the PIP rotation induced by MCP flexion. The total effective angular input to the PIP-side transmission is defined as

\begin{equation}
u_P=-\frac{q_P}{r_P}+\theta_3^{M}.
\label{eq:pip_input_distribution}
\end{equation}

When the distal stopper is engaged and the finger moves without contact, the linkage follows the nominal coupled mode, for which the effective input produces PIP rotation:

\begin{equation}
\theta_3=u_P
=-\frac{q_P}{r_P}+\theta_3^{M}.
\label{eq:pip_angle}
\end{equation}

Unlike the MCP and PIP joints, the DIP joint is not independently actuated. During nominal coupled motion, the mechanical stopper prevents relative rotation between the two spring-connected intermediate links. These links act as one rigid member, and the distal transmission reduces to an equivalent four-bar linkage. The corresponding coupled DIP rotation is defined as

\begin{equation}
\theta_4^{C}=f_{4B}(\theta_3),
\label{eq:dip_coupled}
\end{equation}

\noindent where $f_{4B}(\cdot)$ represents the geometric input--output relationship of the stopper-constrained four-bar linkage. As shown in Fig.~\ref{fig3}(b), the linkage satisfies the planar vector-loop closure condition

\begin{equation}
\begin{aligned}
L_{12}\mathbf{e}(\beta_{12})
+L_{23}\mathbf{e}(\beta_{23})
+L_{34}\mathbf{e}(\beta_{34})\\
=L_{15}\mathbf{e}(\beta_{15})
+L_{54}\mathbf{e}(\beta_{54}),
\end{aligned}
\label{eq:fourbar_closure}
\end{equation}

\noindent where
$\mathbf{e}(\beta)=[\cos\beta,\sin\beta]^T$, $\beta_{ij}$ denotes the orientation angle of the vector from $P_i$ to $P_j$ with respect to the reference line. In the nominal coupled mode, the mechanical stopper maintains the included angle $\alpha$ between $L_{23}$ and $L_{34}$. For the assembly configuration shown in Fig.~\ref{fig3}(b), their orientations satisfy $\beta_{34}=\beta_{23}-(\pi-\alpha)$.

Consequently, $L_{23}$ and $L_{34}$ form an equivalent rigid link with an effective length

\begin{equation}
L_{24}^{C}
=
\sqrt{
L_{23}^{2}+L_{34}^{2}
-2L_{23}L_{34}\cos\alpha
}.
\label{eq:equivalent_link}
\end{equation}

For the design parameters in Table~\ref{tab:design_parameters}, the equivalent link length is approximately 12.00 mm. The corresponding linkage configuration is obtained by satisfying Eq.~\eqref{eq:fourbar_closure} on the assembly branch defined by the initial configuration. The resulting relative DIP rotation determines $f_{4B}(\theta_3)$.

Because the equivalent four-bar linkage is non-parallelogram, its geometric transmission ratio varies with the PIP configuration.

\begin{figure}
    \centering
    \includegraphics[width=0.99\columnwidth]{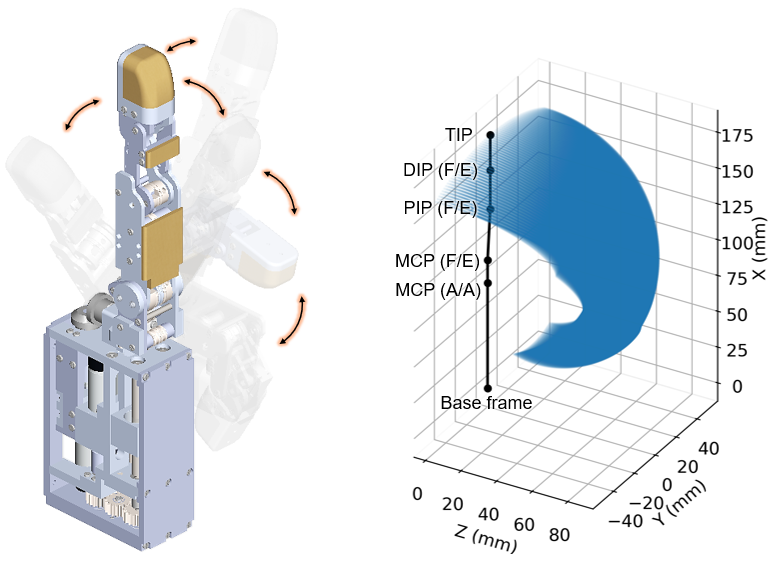}
    \caption{Reachable fingertip workspace during unconstrained motion.}
    \label{fig:workspace}
\end{figure}

The free-motion fingertip workspace shown in Fig.~\ref{fig:workspace} was obtained from the prescribed joint ranges and the nominal coupled PIP--DIP kinematics. The Denavit--Hartenberg parameters of the finger are summarized in Table~\ref{tab:dh_parameters}, the joint ranges used in the subsequent motion and workspace analyses are listed in Table~\ref{tab:joint_rom}, and the kinematic parameters are summarized in Table~\ref{tab:design_parameters}.

\begin{table}[h]
\centering
\caption{Denavit--Hartenberg Parameters of the Finger.}
\label{tab:dh_parameters}
\begin{tabular}{c c c c c}
\toprule
$\boldsymbol{i}$
& $\boldsymbol{a_{i-1}}$
& $\boldsymbol{\alpha_{i-1}}$
& $\boldsymbol{d_i}$
& $\boldsymbol{q_i}$ \\
\midrule
1 & $a_0$ & $0$     & $0$ & $\theta_1$ \\
2 & $a_1$ & $\pi/2$ & $0$ & $\theta_2-\theta_{23}$ \\
3 & $a_2$ & $0$     & $0$ & $\theta_3+\theta_{23}$ \\
4 & $a_3$ & $0$     & $0$ & $\theta_4$ \\
5 & $a_4$ & $0$     & $0$ & $0$ \\
\bottomrule
\end{tabular}
\end{table}

\begin{table}[h]
\centering
\caption{Range of Motion of the Finger Joints.}
\label{tab:joint_rom}
\begin{tabular}{c c c c}
\toprule
$\boldsymbol{\theta_1}$
& $\boldsymbol{\theta_2}$
& $\boldsymbol{\theta_3}$
& $\boldsymbol{\theta_4}$  (Coupled) \\
\midrule
$-\pi/6$ to $\pi/6$
& $-\pi/2$ to $0$
& $-\pi/2$ to $0$
& $-3\pi/20$ to $0$ \\
\bottomrule
\end{tabular}%
\end{table}

\begin{table}[h]
\centering
\caption{Design Parameters of the Proposed Finger.}
\label{tab:design_parameters}
\resizebox{\columnwidth}{!}{%
\begin{tabular}{c l c}
\toprule
\textbf{Symbol} & \textbf{Description} & \textbf{Value} \\
\midrule
$r_M$ 
& Radius of MCP driven pulley 
& 6.00 mm \\

$r_P$ 
& Radius of PIP driven pulley 
& 4.00 mm \\

$L_{12}$ 
& Length of link $L_{12}$ 
& 17.18 mm \\

$L_{23}$ 
& Length of link $L_{23}$ 
& 8.00 mm \\

$L_{34}$ 
& Length of link $L_{34}$ 
& 5.80 mm \\

$L_{54}$ 
& Length of link $L_{54}$ 
& 20.00 mm \\

$L_{15}$ 
& D$^1$ B$^2$ two pivots $P_1$ and $P_5$ 
& 8.00 mm \\

$\theta_{i1}$ 
& Initial angle of link $L_{12}$ RRL$^3$ 
& $17.00^\circ$ \\

$\theta_{i2}$ 
& Initial angle of link $L_{54}$ RRL$^3$ 
& $7.86^\circ$ \\

$\alpha$ 
& IA$^4$ B$^2$ links $L_{23}$ and $L_{34}$ 
& $120.00^\circ$ \\

$a_0$ 
& Link length from F$^5$ 0 to 1 
& 73.5 mm \\

$a_1$ 
& Link length from F$^5$ 1 to 2 
& 16.00 mm \\

$a_2$ 
& Link length from F$^5$ 2 to 3 
& 36.03 mm \\

$a_3$ 
& Link length from F$^5$ 3 to 4 
& 27.33 mm \\

$a_4$ 
& Link length from F$^5$ 4 to 5 
& 24.50 mm \\

$\theta_{23}$ 
& RO$^6$ B$^2$ F$^5$ 2 to 3 
& $2.39^\circ$ \\

$k_s$ 
& Spring constant 
& $0.10~\mathrm{N\!\cdot\!mm/deg}$ \\

$\phi_{s0}$
& Spring preload displacement
& $60^\circ$ \\
\bottomrule
\multicolumn{3}{l}{D$^1$: Distance, B$^2$: Between,}\\
\multicolumn{3}{l}{RRL$^3$: Relative to the Reference Line,}\\
\multicolumn{3}{l}{ IA$^4$: Included angle, F$^6$: Frames,}\\
\multicolumn{3}{l}{RO$^5$: Rotational Offset.}
\end{tabular}%
}
\end{table}

When contact constrains the PIP joint, continued PIP-side actuation permits relative rotation between the two spring-connected intermediate links. Let $\phi_A$ denote the positive relative angular displacement associated with this adaptive motion, measured from the stopper-constrained configuration. As the adaptive motion progresses, the included angle between $L_{23}$ and $L_{34}$ decreases according to

\begin{equation}
\alpha_A=\alpha-\phi_A,
\qquad \phi_A\geq0,
\label{eq:adaptive_link_angle}
\end{equation}

\noindent where $\alpha_A$ is the instantaneous included angle during adaptive motion. Accordingly, the relative orientation of the two links changes, modifying the geometric closure of the distal linkage.

Let $F_{4B}(\theta_3,\alpha_A)$ denote the DIP rotation obtained from the linkage closure constraints for a given PIP configuration and internal included angle, with the contact-constrained linkage configuration taken into account. The additional DIP rotation caused by adaptation is defined as

\begin{equation}
\begin{aligned}
\theta_4^A
&=g_A(\theta_3,\phi_A)\\
&=F_{4B}(\theta_3,\alpha-\phi_A)
-F_{4B}(\theta_3,\alpha),
\end{aligned}
\label{eq:dip_adaptive}
\end{equation}

\noindent where $g_A(\theta_3,0)=0$. The nominal coupled DIP rotation is recovered when $\phi_A=0$, satisfying $f_{4B}(\theta_3)=F_{4B}(\theta_3,\alpha)$. The total DIP rotation is therefore expressed as the sum of the nominal coupled and additional adaptive motions:

\begin{equation}
\theta_4
=\theta_4^C+\theta_4^A
=f_{4B}(\theta_3)+g_A(\theta_3,\phi_A).
\label{eq:dip_total}
\end{equation}

Thus, the additional DIP rotation depends on the PIP configuration at contact and the relative angular displacement of the spring-connected links. For a PIP joint constrained at a contact angle $\theta_{3,c}$, the maximum additional DIP flexion is determined by the allowable relative rotation of the spring-connected links. The maximum adaptive rotation magnitude is expressed as

\begin{equation}
\left|\theta_4^A\right|_{\max}(\theta_{3,c})
=
\max_{\phi_A\in\Omega(\theta_{3,c})}
\left|g_A(\theta_{3,c},\phi_A)\right|,
\label{eq:adaptive_max}
\end{equation}

\noindent where $\Omega(\theta_{3,c})$ denotes the admissible range of the positive adaptive displacement for a given PIP contact angle, subject to the linkage closure constraints and mechanical stops. As $\phi_A$ increases, the included angle $\alpha_A$ decreases, producing additional DIP flexion.

\section{Experimental Validation}

Experiments were conducted to evaluate the coupled and adaptive motion characteristics, fingertip force, and object-contact behavior of the fabricated finger. Motion tests focused on four behaviors: induced by MCP F/E joint, coupled motion of DIP and PIP joints induced by PIP F/E joint, adaptive motion of DIP joint, and MCP A/A motion. The main specifications of the finger are summarized in Table~\ref{tab:specification}.

\begin{table}[h]
\centering
\caption{Technical specifications.}
\label{tab:specification}
\resizebox{\columnwidth}{!}{%
\begin{tabular}{l l}
\toprule
\textbf{Specification} & \textbf{Proposed finger}\\
\midrule
DoF$^1$ & 4 \\
DoA$^2$ & 3 \\
Motion & MCP A/A \& F/E, PIP F/E, DIP F/E \\
Fingertip force & 4.56 N \\
Backdrivable & No \\
Weight &  193.6 g\\
Power supply & 12V DC\\
Communication & CAN \\
Motor PN$^3$ & DCX08M EB KL 12V (Maxon) \\
Gearhead & GPX08 A 16 to 1 for F/E (Maxon) \\
\bottomrule
\multicolumn{2}{l}{DoF$^1$: Degrees of Freedom,}\\
\multicolumn{2}{l}{DoA$^2$: Degrees of Actuation,}\\
\multicolumn{2}{l}{PN$^3$: Part Number.}
\end{tabular}
}
\end{table}

\subsection{Motion Evaluation}

\begin{figure}
    \centering
    \includegraphics[width=0.99\columnwidth]{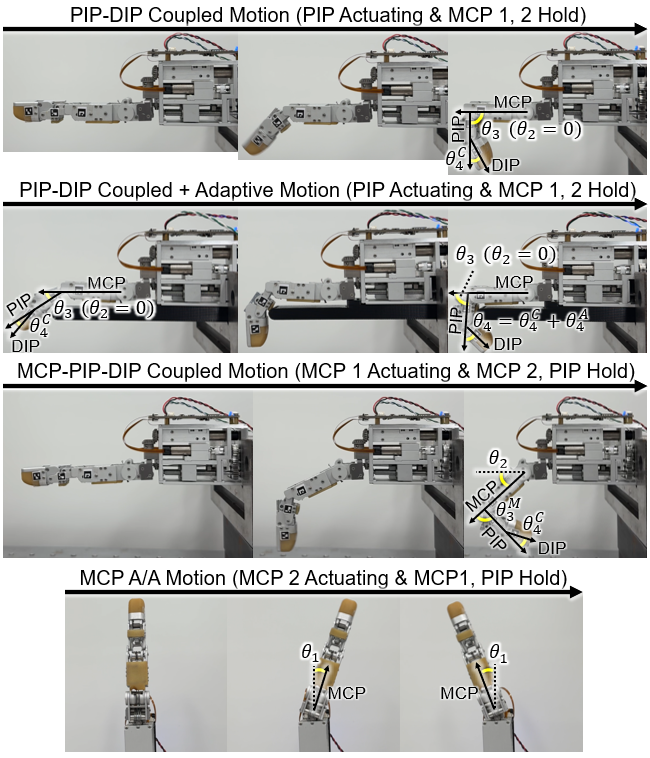}
    \caption{Four types of the motions of fabricated finger.}
    \label{motiontest}
\end{figure}

Four types of motion were evaluated, as shown in Fig.~\ref{motiontest}. For each motion, the joint angles defined in Section~\ref{sec:KA} are indicated in Fig.~\ref{motiontest}, with coupled and adaptive motions distinguished. In the first test, only the motor corresponding to PIP F/E was actuated, while the motors for MCP F/E and A/A remained stationary. This test demonstrated the coupled PIP--DIP motion before the onset of adaptive motion. In the second test, only the PIP F/E motor was actuated, but a jig was placed beneath the finger module to constrain the middle phalanx and induce adaptive motion. Jigs of different lengths were used to demonstrate the resulting adaptive motion under different contact configurations. In the third test, only the MCP F/E motor was actuated, while the PIP F/E and MCP A/A motors remained stationary. This test demonstrated the MCP--PIP coupled motion. Since the PIP and DIP joints are also mechanically coupled, MCP actuation induced coupled motion of both the PIP and DIP joints. In the fourth test, only the MCP A/A motor was actuated for demonstrating independent MCP A/A motion.

In the motion tests, it was difficult to measure the adaptive motion over the entire range of PIP joint angles at which contact could occur. Therefore, ArUco markers with IDs 1--3 were attached to the proximal, middle, and distal phalanges, respectively, to measure the joint configurations from captured images. The relative orientation between IDs 1 and 2 was used to determine the PIP angle, while the relative orientation between IDs 1 and 3 was used to determine the cumulative PIP and DIP rotation. A total of 76 and 111 valid measurements were obtained for the coupled and adaptive motions. For the adaptive motion, the maximum achievable distal flexion was measured at different PIP joint configurations.

\begin{figure}
    \centering
    \includegraphics[width=0.99\columnwidth]{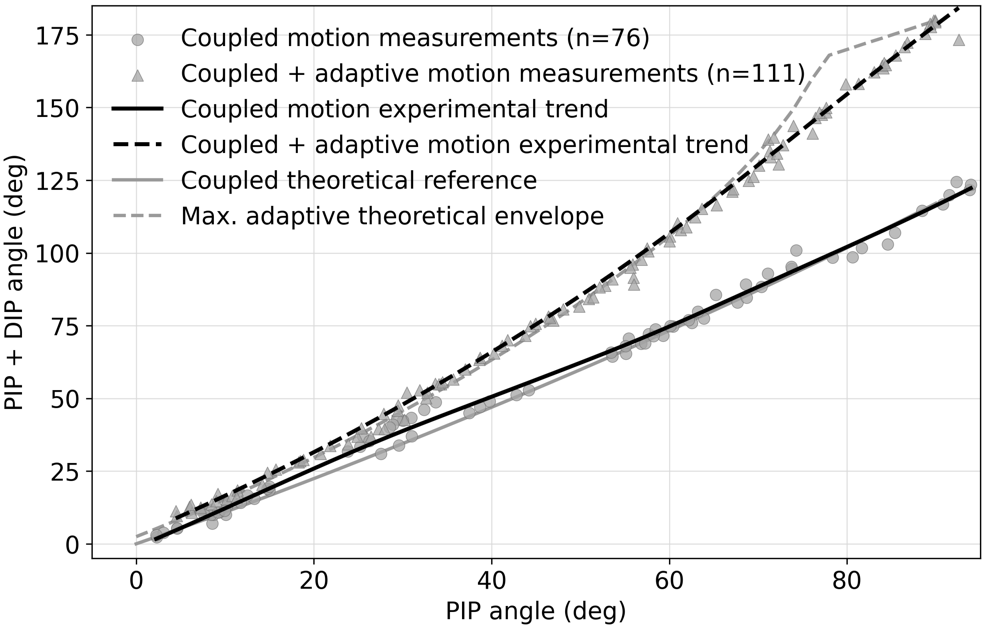}
    \caption{Coupled and adaptive PIP--DIP motion characteristics.}
    \label{coupled_adaptive}
\end{figure}

Figure~\ref{coupled_adaptive} shows the measured coupled and adaptive motions in terms of the PIP angle and the cumulative PIP and DIP flexion. For visualization, the joint flexion angles are presented as positive magnitudes, whereas the kinematic analysis follows the signed joint-angle convention defined in Fig.~\ref{fig3}(a). Accordingly, the horizontal and vertical axes represent $|\theta_3|$ and $|\theta_3|+|\theta_4|$. The circular and triangular markers represent the coupled-motion measurements and maximum coupled-plus-adaptive measurements. The black solid and dashed lines indicate the trends obtained from the measured data for the coupled and maximum coupled-plus-adaptive motions. The gray solid and dashed lines represent the corresponding theoretical results calculated from the four-bar linkage geometry using Eqs.~\eqref{eq:dip_total} and \eqref{eq:adaptive_max}.

The measured coupled motion closely matched the theoretical results. For quantitative comparison, 71 of the 76 coupled-motion measurements were within the theoretical domain. The remaining five measurements were retained in Fig.~\ref{coupled_adaptive} but excluded from the error calculation to avoid extrapolation beyond the theoretical range. The mean absolute error (MAE) and root mean square error (RMSE) were 2.54$^\circ$ and 3.60$^\circ$, respectively, indicating good agreement between the measured coupled motion and the geometric prediction.

For the maximum coupled-plus-adaptive motion, 110 of the 111 valid measurements were within the theoretical domain, while the remaining measurement was excluded from the quantitative comparison. The MAE and RMSE relative to the theoretical maximum envelope were 3.63$^\circ$ and 5.60$^\circ$ with a maximum absolute difference of 18.95$^\circ$. The measured values generally became smaller than the theoretical limits beyond approximately 70$^\circ$ of PIP flexion, before approaching the theoretical values again near the maximum PIP angle. This deviation was partially attributed to mechanical interference between the linkage components connecting the PIP and DIP joints. As the PIP joint continued to rotate, the relative positions of these components changed, reducing the interference and allowing the measured configurations to approach the theoretical limits.

\subsection{Fingertip Force Measurement}

The fingertip force produced by MCP and PIP actuation was evaluated under a fixed joint configuration to characterize the force output of the fabricated finger. A compression/tension load cell (QLMH-25, BIRAN) equipped with a hemispherical cap was used for the measurement. The fingertip was positioned above the hemispherical cap and actuated to press against it. Because a single-axis load cell was used, the measured force represents the component acting along the measurement axis. The load-cell readings were converted to force using $g=9.812~\mathrm{m/s^2}$. The data were sampled at approximately 80 Hz, and the experiment was repeated five times under the same test conditions.

\begin{figure}
    \centering
    \includegraphics[width=0.99\columnwidth]{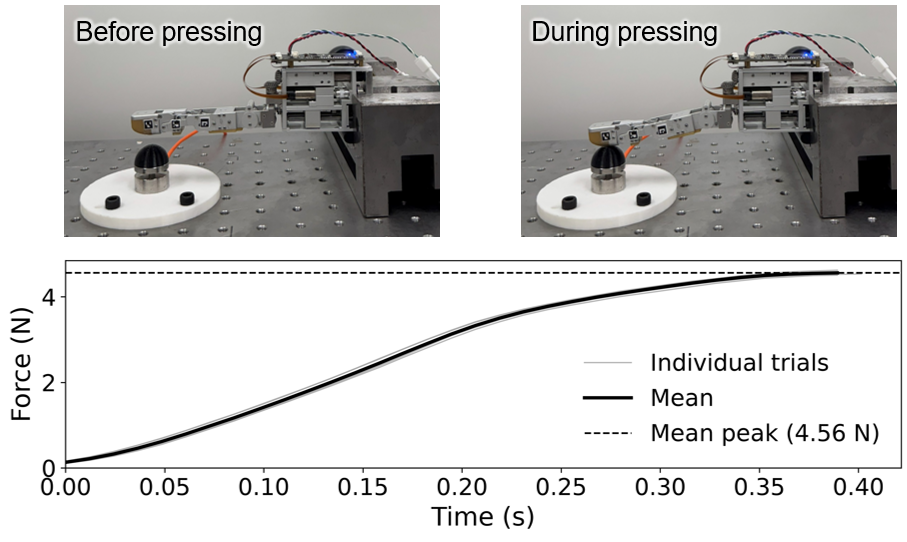}
    \caption{Experimental setup and measurement results of fingertip force.}
    \label{tipforce}
\end{figure}

Figure~\ref{tipforce} shows the measured force profiles from the five trials and their mean. The gray lines represent the individual trials, the black solid line denotes the mean force profile, and the black dashed line indicates the mean peak force. The measured force increased during actuation and approached a nearly constant level toward the end of the measurement interval. The mean peak force across the five trials was 4.56 N, with a sample standard deviation of 0.045 N. The individual peak forces ranged from 4.50 to 4.62 N, corresponding to a coefficient of variation of approximately 0.98\%. These results indicate consistent peak force output under the tested configuration.

The fingertip force is influenced by the joint configuration and mechanical transmission characteristics of the finger. In particular, the configuration-dependent linkage geometry affects the relationship between the transmitted joint torque and the resulting fingertip force. Therefore, the measured peak force characterizes the tested configuration rather than the maximum force available over the entire joint range.

\subsection{Object Contact Experiments}

Object-contact experiments were conducted to examine the ability of the fabricated finger to establish different joint configurations in response to object geometry. Seven objects were considered, including four cylindrical objects with diameters of 7.9, 13.8, 46.4, and 90.2 mm, two cuboid objects with cross-sectional dimensions of 13.5 × 13.5 mm and 59.2 × 28.0 mm, and one irregularly shaped object.

\begin{figure}
    \centering
    \includegraphics[width=0.99\columnwidth]{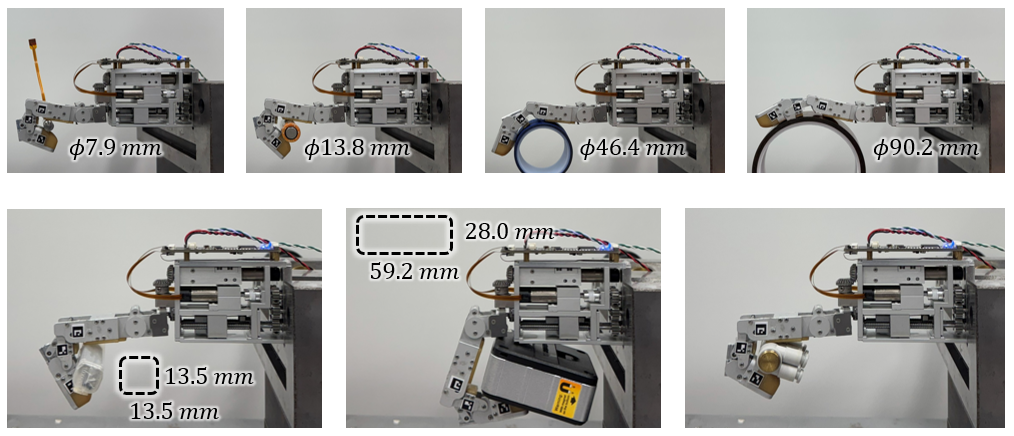}
    \caption{Postures of the finger during interaction with various objects.}
    \label{object_contact}
\end{figure}

Figure~\ref{object_contact} shows the resulting finger configurations during interaction with these objects. The PIP and DIP joint postures varied according to the dimensions and external geometries of the contacted objects. These demonstrations illustrate the range of contact configurations achievable by the proposed finger with its coupled–adaptive distal mechanism.

\section{Conclusion}

This work presented a hybrid tendon--linkage robotic finger with coupled and adaptive joint motions. The finger provides four DoFs using three actuators through a bevel-gear transmission for MCP A/A, bidirectional tendon transmissions for MCP and PIP F/E, and a spring-loaded four-bar linkage for the non-actuated DIP joint. The kinematic analysis described the MCP--PIP coupling induced by tendon routing and the nonlinear PIP--DIP coupling determined by the four-bar linkage. Experiments with the fabricated finger demonstrated four motion types. ArUco-based measurements showed good agreement with the theoretical coupled-motion prediction, while the maximum adaptive motion exhibited configuration-dependent deviations due to internal linkage interference. The fabricated finger also achieved different contact postures with cylindrical, cuboid, and irregularly shaped objects. The mean peak fingertip force was measured.

The proposed finger has several limitations. The maximum adaptive motion could not be achieved in certain PIP configurations due to mechanical interference between internal linkage components. Although the PIP joint was backdrivable, the MCP joint was not, resulting in limited overall backdrivability of the finger. Therefore, future work will focus on improving the transmission efficiency and adaptive motion of the finger. The tendon-routing paths will be simplified to reduce friction and improve backdrivability, and the transmission components and motor reduction ratios will be modified to increase fingertip force. The distal linkage structure will also be revised to reduce interference between components and improve the available adaptive motion range. In addition, the DoFs and linkage structure will be modified to develop an opposable thumb, allowing the proposed finger mechanism to be extended to a multi-fingered robotic hand.

\bibliographystyle{IEEEtran}
\bibliography{bibtex}

@article{laliberte1998simulation,
  title={Simulation and design of underactuated mechanical hands},
  author={Lalibert{\'e}, Thierry and Gosselin, Clement M},
  journal={Mechanism and machine theory},
  volume={33},
  number={1-2},
  pages={39--57},
  year={1998},
  publisher={Elsevier}
}

@article{tavakoli2015underactuated,
  title={Underactuated anthropomorphic hands: Actuation strategies for a better functionality},
  author={Tavakoli, Mahmoud and Enes, Baptiste and Santos, Joana and Marques, Lino and de Almeida, Anibal T},
  journal={Robotics and Autonomous Systems},
  volume={74},
  pages={267--282},
  year={2015},
  publisher={Elsevier}
}

@article{inouye2012novel,
  title={A novel synthesis of computational approaches enables optimization of grasp quality of tendon-driven hands},
  author={Inouye, Joshua M and Kutch, Jason J and Valero-Cuevas, Francisco J},
  journal={IEEE transactions on robotics},
  volume={28},
  number={4},
  pages={958--966},
  year={2012},
  publisher={IEEE}
}

@article{kurita2011human,
  title={Human-sized anthropomorphic robot hand with detachable mechanism at the wrist},
  author={Kurita, Yuichi and Ono, Yasuhiro and Ikeda, Atsutoshi and Ogasawara, Tsukasa},
  journal={Mechanism and Machine Theory},
  volume={46},
  number={1},
  pages={53--66},
  year={2011},
  publisher={Elsevier}
}

@article{sun2012novel,
  title={A novel coupled and self-adaptive under-actuated multi-fingered hand with gear--rack--slider mechanism},
  author={Sun, Jie and Zhang, Wenzeng},
  journal={Journal of Manufacturing Systems},
  volume={31},
  number={1},
  pages={42--49},
  year={2012},
  publisher={Elsevier}
}

@article{zhang2013two,
  title={Two-DOF coupled and self-adaptive (COSA) finger: a novel under-actuated mechanism},
  author={Zhang, Wenzeng and Zhao, Deyang and Zhou, Haipeng and Sun, Zhenguo and Du, Dong and Chen, Qiang},
  journal={International Journal of Humanoid Robotics},
  volume={10},
  number={02},
  pages={1330001},
  year={2013},
  publisher={World Scientific}
}

@article{li2012development,
  title={Development of multi-fingered robotic hand with coupled and directly self-adaptive grasp},
  author={Li, Guoxuan and Liu, Han and Zhang, Wenzeng},
  journal={International Journal of Humanoid Robotics},
  volume={9},
  number={04},
  pages={1250034},
  year={2012},
  publisher={World Scientific}
}

@article{liu2019coupled,
  title={A coupled and indirectly self-adaptive under-actuated hand with double-linkage-slider mechanism},
  author={Liu, Siyun and Zhang, Wenzeng and Sun, Jie},
  journal={Industrial Robot: the international journal of robotics research and application},
  volume={46},
  number={5},
  pages={660--671},
  year={2019},
  publisher={Emerald Publishing Limited}
}

@article{zhang2023design,
  title={Design of a multi-mode mechanical finger based on linkage and tendon fusion transmission},
  author={Zhang, Yi and Zhao, Qian and Deng, Hua and Xu, Xiaolei},
  journal={Biomimetics},
  volume={8},
  number={3},
  pages={316},
  year={2023},
  publisher={MDPI}
}

@article{kim2025dual,
  title={Dual-tendon routing: Tendon routing for under-actuated tendon-driven soft hand-wearable robot},
  author={Kim, Byungchul and Jeong, Useok and Cho, Kyu-Jin},
  journal={IEEE Robotics and Automation Letters},
  volume={10},
  number={4},
  pages={3612--3619},
  year={2025},
  publisher={IEEE}
}

@inproceedings{li2010study,
  title={Study on coupled and self-adaptive finger for robot hand with parallel rack and belt mechanisms},
  author={Li, Guoxuan and Zhang, Wenzeng},
  booktitle={2010 IEEE International Conference on Robotics and Biomimetics},
  pages={1110--1115},
  year={2010},
  organization={IEEE}
}

@article{li2014coupled,
  title={Coupled and self-adaptive under-actuated finger with a novel s-coupled and secondly self-adaptive mechanism},
  author={Li, Guoxuan and Zhang, Chi and Zhang, Wenzeng and Sun, Zhenguo and Chen, Qiang},
  journal={Journal of Mechanisms and Robotics},
  volume={6},
  number={4},
  pages={041010},
  year={2014},
  publisher={American Society of Mechanical Engineers}
}

@article{lee2023design,
  title={Design of hybrid fully actuated and self-adaptive mechanism for anthropomorphic robotic finger},
  author={Lee, Chun-Tse and Chang, Jen-Yuan},
  journal={Journal of Mechanisms and Robotics},
  volume={15},
  number={4},
  pages={041004},
  year={2023},
  publisher={American Society of Mechanical Engineers}
}

@article{cheng2025design,
  title={Design and Optimization of an Anthropomorphic Robot Finger},
  author={Cheng, Ming and Jiang, Li and Liu, Ziqi},
  journal={Biomimetics},
  volume={10},
  number={3},
  pages={170},
  year={2025},
  publisher={MDPI}
}

@article{min2025antagonistic,
  title={Antagonistic Bowden-Cable Actuation of a Lightweight Robotic Hand: Toward Dexterous Manipulation for Payload Constrained Humanoids},
  author={Min, Sungjae and Kim, Hyungjoo and Shim, David Hyunchul},
  journal={arXiv preprint arXiv:2512.24657},
  year={2025}
}

@article{li2026design,
  title={Design and evaluation of a tendon-and-linkage hybrid-driven humanoid dexterous hand},
  author={Li, Ke and Meng, Fanlin and Liu, Linjie and Nataraj, Raviraj and Sun, Ning and Wei, Na},
  journal={Scientific Reports},
  year={2026},
  publisher={Nature Publishing Group UK London}
}

@inproceedings{yoon2016underactuated,
  title={Underactuated finger mechanism for natural motion and self-adaptive grasping towards bionic partial hand},
  author={Yoon, Dukchan and Lee, Geon and Lee, Sungon and Choi, Youngjin},
  booktitle={2016 6th IEEE International Conference on Biomedical Robotics and Biomechatronics (BioRob)},
  pages={548--553},
  year={2016},
  organization={IEEE}
}

@article{su2023sau,
  title={SAU-RFC hand: A novel self-adaptive underactuated robot hand with rigid-flexible coupling fingers},
  author={Su, Congjia and Wang, Rui and Lu, Tao and Wang, Shuo},
  journal={Robotica},
  volume={41},
  number={2},
  pages={511--529},
  year={2023},
  publisher={Cambridge University Press}
}

@article{birglen2004kinetostatic,
  title={Kinetostatic analysis of underactuated fingers},
  author={Birglen, Lionel and Gosselin, Cl{\'e}ment M},
  journal={IEEE Transactions on Robotics and Automation},
  volume={20},
  number={2},
  pages={211--221},
  year={2004},
  publisher={IEEE}
}

@article{birglen2010flapping,
  title={From flapping wings to underactuated fingers and beyond: a broad look to self-adaptive mechanisms},
  author={Birglen, Lionel},
  journal={Mechanical Sciences},
  volume={1},
  number={1},
  pages={5--12},
  year={2010},
  publisher={Copernicus Publications G{\"o}ttingen, Germany}
}

@inproceedings{aukes2011varying,
  title={Varying spring preloads to select grasp strategies in an adaptive hand},
  author={Aukes, Daniel and Heyneman, Barrett and Duchaine, Vincent and Cutkosky, Mark R},
  booktitle={2011 IEEE/RSJ International conference on intelligent robots and systems},
  pages={1373--1379},
  year={2011},
  organization={IEEE}
}

@article{qiao2019self,
  title={Self-adaptive grasp process and equilibrium configuration analysis of a 3-DOF UACT robotic finger},
  author={Qiao, Shangling and Guo, Hongwei and Liu, Rongqiang and Deng, Zongquan},
  journal={Mechanism and Machine Theory},
  volume={133},
  pages={250--266},
  year={2019},
  publisher={Elsevier}
}

@article{li2014indirectly,
  title={Indirectly self-adaptive underactuated robot hand with block-linkage mechanisms},
  author={Li, Guoxuan and Jin, Jian and Deschamps-Berger, Sylvain and Sun, Zhenguo and Zhang, Wenzeng and Chen, Qiang},
  journal={International Journal of Precision Engineering and Manufacturing},
  volume={15},
  number={8},
  pages={1553--1562},
  year={2014},
  publisher={Springer}
}

@article{c1,
title = {Robotic hand: A review on linkage-driven finger mechanisms of prosthetic hands and evaluation of the performance criteria},
journal = {Mechanism and Machine Theory},
volume = {145},
pages = {103677},
year = {2020},
issn = {0094-114X},
author = {S. Reza Kashef and Samane Amini and Alireza Akbarzadeh}
}

@article{c9,
  title={A comprehensive review of dexterous robotic hands: Design, implementation, and evaluation},
  author={Lin, Yu-Ying and Raj, Ritik and Juang, Jia-Yang},
  journal={Bioinspiration \& Biomimetics},
  volume={20},
  number={4},
  pages={041003},
  year={2025},
  publisher={IOP Publishing}
}

@inproceedings{c15,
  title={A tendon-driven dexterous hand design with tactile sensor array for grasping and manipulation},
  author={Fang, Bin and Sun, Fuchun and Chen, Yang and Zhu, Chang and Xia, Ziwei and Yang, Yiyong},
  booktitle={2019 IEEE International Conference on Robotics and Biomimetics (ROBIO)},
  pages={203--210},
  year={2019},
  organization={IEEE}
}

@article{c14,
  title={Design principle of a dual-actuated robotic hand with anthropomorphic self-adaptive grasping and dexterous manipulation abilities},
  author={Sun, Bai-Yang and Gong, Xuan and Liang, Jiejunyi and Chen, Wen-Bin and Xie, Zhi-Liang and Liu, Chuang and Xiong, Cai-Hua},
  journal={IEEE Transactions on Robotics},
  volume={38},
  number={4},
  pages={2322--2340},
  year={2021},
  publisher={IEEE}
}

@article{c20,
  title={A low-cost linkage-spring-tendon-integrated compliant anthropomorphic robotic hand: MCR-Hand III},
  author={Yang, Haosen and Wei, Guowu and Ren, Lei and Qian, Zhihui and Wang, Kunyang and Xiu, Haohua and Liang, Wei},
  journal={Mechanism and Machine Theory},
  volume={158},
  pages={104210},
  year={2021},
  publisher={Elsevier}
}

@article{c88,
  title={Design of a reconfigurable three-finger gripper for versatile grasping tasks},
  author={Kang, HyoJae and Han, Daehee and Son, KunWoo and Kang, Min-Sung},
  journal={Robotics and Autonomous Systems},
  volume={196},
  pages={105264},
  year={2026},
  publisher={Elsevier}
}

@inproceedings{c90,
  title={Anthropomorphic robot hand: Gifu hand III},
  author={Mouri, Tetsuya and Kawasaki, Haruhisa and Yoshikawa, Keisuke and Takai, Jun and Ito, Satoshi and others},
  booktitle={Proc. Int. Conf. ICCAS},
  pages={1288--1293},
  year={2002}
}

@article{c91,
title = {An open-source anthropomorphic robot hand system: HRI hand},
journal = {HardwareX},
volume = {7},
pages = {e00100},
year = {2020},
issn = {2468-0672},
author = {Hyeonjun Park and Donghan Kim}
}

@misc{c92,
  author       = {{Inspire Robots Technology Co., Ltd.}},
  title        = {{RH56BFX}},
  howpublished = {\url{https://en.inspire-robots.com/product/rh56bfx}},
  note         = {Accessed: Nov. 09, 2026}
}

@misc{c95,
  author       = {{Shadow Robot Company}},
  title        = {{Shadow Dexterous Hand Series}},
  howpublished = {\url{https://shadowrobot.com/dexterous-hand-series/}},
  note         = {Accessed: Nov. 09, 2026}
}

@article{c99,
  title={Design of a Fully Actuated 4-DOF Robotic Finger With Joint-Specific Hybrid Remote Actuation},
  author={Kang, Hyojae and Jung, Hyun-mok and Lee, Joonho and Park, Dongil and Do, Hyunmin and Park, Jongwoo and Ahn, Jeongdo},
  journal={arXiv preprint arXiv:2610.10180},
  year={2026}
}

\end{document}